\documentclass[runningheads]{llncs}
\usepackage{graphicx}
\usepackage{amssymb}
\usepackage{color}
\usepackage{url}
\usepackage{cite}
\usepackage[normalem]{ulem}

\usepackage{caption}

\begin{document}
\title{Learnable Cross-modal Knowledge Distillation for Multi-modal Learning with Missing Modality}

\author{Hu Wang\inst{1} \and
Congbo Ma\inst{1} \and
Jianpeng Zhang\inst{2} \and
Yuan Zhang\inst{1} \and
Jodie Avery\inst{1} \and
Louise Hull\inst{1} \and
Gustavo Carneiro\inst{3}
}
\authorrunning{H. Wang et al.}
%
\titlerunning{Learnable Cross-modal Knowledge Distillation}

\institute{The University of Adelaide, Australia \and
Alibaba DAMO Academy, China \and
Centre for Vision, Speech and Signal Processing, University of Surrey, UK
}
\maketitle              
\begin{abstract}

The problem of missing modalities is both critical and non-trivial to be handled in multi-modal models. It is common for multi-modal tasks that certain modalities contribute more compared to other modalities, and if those important modalities are missing, the model performance drops significantly. Such fact remains unexplored by current multi-modal approaches that recover the representation from missing modalities by feature reconstruction or blind feature aggregation from other modalities, instead of extracting useful information from the best performing modalities. In this paper, we propose a Learnable Cross-modal Knowledge Distillation (LCKD) model to adaptively identify important modalities and distil knowledge from them to help other modalities from the cross-modal perspective for solving the missing modality issue. Our approach introduces a teacher election procedure to select the most ``qualified'' teachers based on their single modality performance on certain tasks. Then, cross-modal knowledge distillation is performed between teacher and student modalities for each task to push the model parameters to a point that is beneficial for all tasks. Hence, even if the teacher modalities for certain tasks are missing during testing, the available student modalities can accomplish the task well enough based on the learned knowledge from their automatically elected teacher modalities. Experiments on the Brain Tumour Segmentation Dataset 2018 (BraTS2018) shows that LCKD outperforms other methods by a considerable margin, improving the state-of-the-art performance by 3.61\% for enhancing tumour, 5.99\% for tumour core, and 3.76\% for whole tumour in terms of segmentation Dice score.

\keywords{Missing modality issue \and Multi-modal learning \and Learnable cross-modal knowledge distillation.}
\end{abstract}

\vspace{-.2cm}
\section{Introduction}
\label{sec:intro}
\vspace{-.2cm}

Multi-modal learning has become a popular research area in computer vision and medical image analysis, with modalities spanning across various media types, including texts, audio, images, videos and multiple sensor data. This approach has been utilised in Robot Control~\cite{noda2014multimodal,wang2020soft}, Visual Question Answering~\cite{kazemi2017show} and Audio-Visual Speech Recognition~\cite{huang2013audio}, as well as in the medical field to improve diagnostic system performance~\cite{dou2020unpaired,wang2022uncertainty}. 
For instance, Magnetic Resonance Imaging (MRI) is a common tool for brain tumour detection that relies on multiple modalities (Flair, T1, T1 contrast-enhanced known as T1c, and T2) rather than a single type of MRI images. 
However, most existing multi-modal methods require complete modalities during training and testing, which limits their applicability in real-world scenarios, where subsets of modalities may be missing during training and testing.

The missing modality issue is a significant challenge in the multi-modal domain, and it has motivated the community to develop approaches that attempt to address this problem. 
Havaei et al.~\cite{havaei2016hemis} developed HeMIS, a model that handles missing modalities using statistical features as embeddings for the model decoding process. Taking one step ahead, Dorent et al.~\cite{dorent2019hetero} proposed an extension to HeMIS via a multi-modal variational auto-encoder (MVAE) to make predictions based on learned statistical features. 
In fact, variational auto-encoder (VAE) has been adopted to generate data from other modalities in the image or feature domains~\cite{chartsias2017multimodal,jing2020incomplete}. 
Yin et al.~\cite{yin2017unified} aimed to learn a unified subspace for incomplete and unlabelled multi-view data. Chen et al.~\cite{chen2019robust} proposed a feature disentanglement and gated fusion framework to separate modality-robust and modality-sensitive features. Ding et al.~\cite{ding2021rfnet} proposed an RFM module to fuse the modal features based on the sensitivity of each modality to different tumor regions and a segmentation-based regularizer to address the imbalanced training problem. Zhang et al.~\cite{zhang2021modality} proposed an MA module to ensure that modality-specific models are interconnected and calibrated with attention weights for adaptive information exchange. Recently, Zhang et al.~\cite{zhang2022mmformer} introduced a vision transformer architecture, MMFormer, that fuses features from all modalities into a set of comprehensive features. 
There are several existing works~\cite{hu2020knowledge,shen2019brain,wang2021acn} proposed to approximate the features from full modalities when one or more modalities are absent. But none work performs cross-modal knowledge distillation. 
From an other point of view, Wang et al.~\cite{wang2021acn} introduced a dedicated training strategy that separately trains a series of models specifically for each missing situation, which requires significantly more computation resources compared with a non-dedicated training strategy. 
An interesting fact about multi-modal problems is that there is always one modality that contributes much more than other modalities for a certain task.
For instance, for brain tumour segmentation, it is known from domain knowledge that T1c scans clearly display the enhanced tumour, but not edema~\cite{chen2019robust}. 
If the knowledge of these modalities can be successfully preserved, the model can produce promising results even when these best performing modalities are not available.
However, the aforementioned methods neglect the contribution biases of different modalities and failed to consider keeping that knowledge.

Aiming at this issue, we propose the non-dedicated training model\footnote{We train one model to handle all of the different missing modality situations.} \underline{\textbf{L}}earnable \underline{\textbf{C}}ross-modal \underline{\textbf{K}}nowledge \underline{\textbf{D}}istillation (LCKD) for tackling the missing modality issue. 
LCKD is able to handle missing modalities in both training and testing by automatically identifying important modalities and distilling knowledge from them to learn the parameters that are beneficial for all tasks while training for other modalities (e.g., there are four modalities and three tasks for the three types of tumours in BraTS2018). 
Our main contributions are:
\begin{itemize}
\item We propose the Learnable Cross-modal Knowledge Distillation (LCKD) model to address missing modality problem in multi-modal learning. 
It is a simple yet effective model designed from the viewpoint of distilling cross-modal knowledge to maximise the performance for all tasks;
\item The LCKD approach is designed to automatically identify the important modalities per task, which helps the cross-modal knowledge distillation process. It also can handle missing modality during both training and testing.
\end{itemize}
The experiments are conducted on the Brain Tumour Segmentation benchmark BraTS2018~\cite{menze2014multimodal,bakas2018identifying}, showing that our LCKD model achieves state-of-the-art performance. In comparison to recently proposed competing methods on BraTS2018, our model demonstrates  better performance in segmentation Dice score by 3.61\% for enhancing tumour, 5.99\% for tumour core, and 3.76\% for whole tumour, on average.

\vspace{-.2cm}
\section{Methodology}
\subsection{Overall Architecture}
\vspace{-.2cm}

\begin{figure}[t]
\centering
\includegraphics[width=0.9\textwidth]{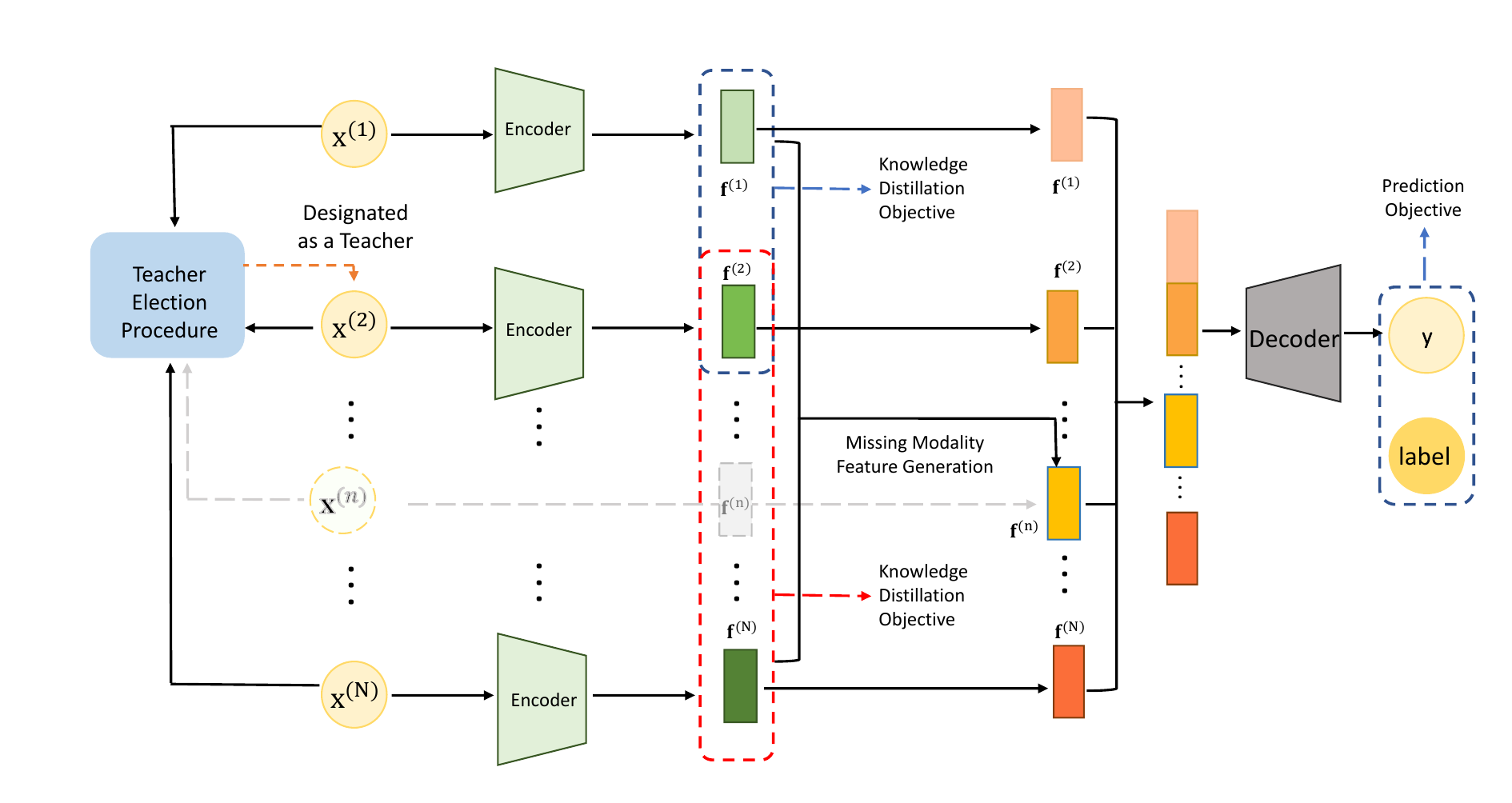}
\vspace{-5mm}
\caption{LCKD model framework for training and testing. The N modalities $\{{x^{(i)}}_{i=1}^{N}\}$ are processed by the encoder to produce the features $\{{f^{(i)}}_{i=1}^{N}\}$, which are concatenated and used by the decoder to produce the segmentation. The teacher is elected using a validation process that selects the top-performing modalities as teachers. Cross-modal distillation is performed by approximating the available students' features to the available teachers' features. Features from missing modalities are generated by averaging the other modalities' features.
}
\label{fig:framework}
\end{figure}

Let us represent the $N$-modality data with $\mathcal{M}_l = \{\mathbf{x}^{(i)}_{l} \}_{i=1}^{N}$, where $\mathbf{x}^{(i)}_{l} \in \mathcal{X}$ denotes the $l^{th}$ data sample and the superscript $^{(i)}$ indexes the modality. To simplify the notation, we omit the subscript $l$ when that information is clear from the context. The label for each set $\mathcal{M}$ is represented by $\mathbf{y} \in \mathcal{Y}$, where $\mathcal{Y}$ represents the ground-truth annotation space. The framework of LCKD is shown in Fig.~\ref{fig:framework}.

Multi-modal segmentation is composed not only of multiple modalities, but also of multiple tasks, such as the three types of tumours in BraTS2018 dataset that represent the three tasks. Take one of the tasks for example. Our model undergoes an external Teacher Election Procedure prior to processing all modalities $\{\mathbf{x}^{(i)}\}_{i=1}^{N} \in \mathcal{M}$ in order to select the modalities that exhibit promising performance as teachers. 
This is illustrated in Fig.~\ref{fig:framework}, where one of the modalities, $\mathbf{x}^{(2)}$, is selected as a teacher, $\{\mathbf{x}^{(1)},\mathbf{x}^{(3)},...,\mathbf{x}^{(N)}\}$ are the students, and $\mathbf{x}^{(n)}$ (with $n \ne 2$) is assumed to be absent. Subsequently, the modalities are encoded to output features $\{\mathbf{f}^{(i)}\}_{i=1}^{N}$, individually. For the modalities that are available, namely  $\mathbf{x}^{(1)}, ..., \mathbf{x}^{(n-1)}, \mathbf{x}^{(n+1)}, ..., \mathbf{x}^{(N)}$, knowledge distillation is carried out between each pair of teacher and student modalities. However, for the absent modality $\mathbf{x}^{(n)}$, its features $\mathbf{f}^{(n)}$ are produced through a missing modality feature generation process from the available features $\mathbf{f}^{(1)}, ..., \mathbf{f}^{(n-1)}, \mathbf{f}^{(n+1)}, ..., \mathbf{f}^{(N)}$.

In the next sections, we explain each module of the proposed Learnable Cross-modal Knowledge Distillation model training and testing with full and missing modalities.

\vspace{-.2cm}
\subsection{Teacher Election Procedure}
\vspace{-.2cm}

Usually, one of the modalities is more useful than others for a certain task, e.g. for brain tumour segmentation, T1c scan clearly displays the enhanced tumour, but it does not clearly show edema~\cite{chen2019robust}. 
Following knowledge distillation (KD)~\cite{hu2020knowledge}, we propose to transfer the knowledge from modalities with promising performance (known as teachers) to other modalities (known as students). The teacher election procedure is further introduced to automatically elect proper teachers for different tasks.

More specifically, in the teacher election procedure, a validation process is applied: for each task $k$ (for $k \in \{1,...,K\}$), the modality with the best performance is selected as the teacher $\mathbf{t}^{(k)}$. Formally, we have:
\begin{equation} \small
    \mathbf{t}^{(k)} = \mathop{\arg\max}\limits_{i \in \{1,...,N\}} \sum_{l=1}^{L}d(F(\mathbf{x}_l^{(i)}; \mathbf{\Theta}), \mathbf{y}_l),
    \label{eq:teacher_elect1}
\end{equation}
where $i$ indexes different modalities,  $F(\cdot; \mathbf{\Theta})$ is the LCKD segmentation model parameterised by $\mathbf{\Theta}$, including the encoder and decoder parameters $\{\theta^{enc}, \theta^{dec}\} \in \mathbf{\Theta}$, and $d(\cdot, \cdot)$ is the function to calculate the Dice score. 
Based on the elected teachers for different tasks, a list of unique teachers (i.e., repetitions are not allowed in the list, so for BraTS, \{T1c, T1c, Flair\} would be reduced to \{T1c, Flair\}) are generated with:
\begin{equation} \small
    \mathbf{T} = \phi(\mathbf{t}^{(1)}, \mathbf{t}^{(2)}, ..., \mathbf{t}^{(k)}, ..., \mathbf{t}^{(K)}),
    \label{eq:teacher_elect2}
\end{equation}
where $\phi$ is the function that returns the unique elements from a given list, and $\mathbf{T} \subseteq \{1,...,N\}$ is the teacher set.

\vspace{-.2cm}
\subsection{Cross-modal Knowledge Distillation}
\vspace{-.2cm}

As shown in Fig. \ref{fig:framework}, after each modality $\mathbf{x}^{(i)}$ is inputted into the encoder parameterised by $\theta^{enc}$, the features $\mathbf{f}^{(i)}$ for each modality is fetched, as in:
\begin{equation} \small
\label{eq:enc}
        \mathbf{f}^{(i)} = f_{\theta^{enc}}(\mathbf{x}^{(i)}).
\end{equation}

The cross-modal knowledge distillation (CKD) is defined by a loss function that approximates all available modalities' features to the available teacher modalities in a pairwise manner for all tasks, as follows:
\begin{equation} \small
\label{eq:kd}
    \ell_{ckd}(\mathcal{D};\theta^{enc}) = \sum_{i \in \mathbf{T};i,j \notin \mathbf{m}}^{N} \| \mathbf{f}^{(i)} - \mathbf{f}^{(j)} \|_p,
\end{equation}
where $\|\cdot\|_p$ presents the p-norm operation, and here we expended the notation of missing modalities to make it more general by assuming a set of modalities $\mathbf{m}$ is missing.
The minimisation of this loss pushes the model parameter values to a point in the parameter space that can maximise the performance of all tasks for all modalities.

\vspace{-.2cm}
\subsection{Missing Modality Feature Generation}
\vspace{-.2cm}

Because of the knowledge distillation between each pair of teachers and students, the features of modalities in the feature space ought to be close to the ``genuine'' features that can uniformly perform well for different tasks. Still assuming that modality set $\mathbf{m}$ is missing, the missing features $\mathbf{f}^{(n)}$ can thus be generated from the available features:
\begin{equation} \small
    \mathbf{f}^{(n)} = \frac{1}{N-|\mathbf{m}|}\sum_{i=1;i \notin \mathbf{m}}^{N} \mathbf{f}^{(i)},
    \label{eq:ftgen}
\end{equation}
where $|\mathbf{m}|$ denotes the number of missing modalities.

\vspace{-.2cm}
\subsection{Training and Testing}
\vspace{-.2cm}

All features encoded from~Eq.~\ref{eq:enc} or generated from~Eq.~\ref{eq:ftgen} are then  concatenated to be fed into the decoder parameterised by $\theta^{dec}$ for predicting
\begin{equation} \small
    \tilde{\mathbf{y}} = f_{\theta^{dec}}(\mathbf{f}^{(1)},...,\mathbf{f}^{(N)}),
    \label{eq:decoder}
\end{equation}
where $\tilde{\mathbf{y}} \in \mathcal{Y}$ is the prediction of the task. 

The training of the whole model is achieved by minimising the following objective function:
\begin{equation} \small
\ell_{\mathit{tot}}(\mathcal{D},\mathbf{\Theta}) = \ell_{\mathit{task}}(\mathcal{D}, \theta^{enc}, \theta^{dec}) + \alpha \ell_{ckd}(\mathcal{D};\theta^{enc}),
\label{eq:tot_loss}
\end{equation}
where $\ell_{\mathit{task}}(\mathcal{D}, \theta^{enc}, \theta^{dec})$ is the objective function for the whole task (e.g., Cross-Entropy and Dice losses are adopted for brain tumour segmentation), and 
$\alpha$ is the trade-off factor between the task objective and cross-modal KD objective.

Testing is based on taking all image modalities available in the input to produce the features from Eq.~\ref{eq:enc}, and generating the features from the missing modalities with~Eq.~\ref{eq:ftgen}, which are then provided to the decoder to predict the segmentation with~Eq.~\ref{eq:decoder}.

\vspace{-.2cm}
\section{Experiments}
\vspace{-.2cm}

\subsection{Data and Implementation Details}
\vspace{-.2cm}

Our model and competing methods are evaluated on the BraTS2018 Segmentation Challenge dataset~\cite{menze2014multimodal,bakas2018identifying}. The task involves segmentation of three sub-regions of brain tumours, namely enhancing tumour (ET), tumour core (TC), and whole tumour (WT). The dataset consists of 3D multi-modal brain MRIs, including Flair, T1, T1 contrast-enhanced (T1c), and T2, with ground-truth annotations. The dataset comprises 285 cases for training, and 66 cases for evaluation. The ground-truth annotations for the training set are publicly available, while the validation set annotations are hidden\footnote{Online evaluation is required at \url{https://ipp.cbica.upenn.edu/}.}.

3D UNet architecture (with 3D convolution and normalisation) is adopted as our backbone network, where the CKD process occurs at the bottom stage of the UNet structure. To optimise our model, we adopt a stochastic gradient descent optimiser with Nesterov momentum~\cite{botev2017nesterov} set to 0.99. L1 loss is adopted for $\ell_{ckd}(.)$ in Eq.~\ref{eq:kd}. 
Batch-size is set to 2. The learning rate is initially set to $10^{-2}$ and gradually decreased via the cosine annealing~\cite{loshchilov2016sgdr} strategy. We trained the LCKD model for 115,000 iterations and use 20\% of the training data as the validation task for teacher election. To simulate modality-missing situations with non-dedicated training of models, we randomly dropped 0 to 3 modalities for each iteration. Our training time is 70.12 hours and testing time is 6.43 seconds per case on one Nvidia 3090 GPU. 19795 MiB GPU memory is used for model training with batch-size 2 and 3789 MiB GPU memory is consumed for model testing with batch-size 1.

\vspace{-.2cm}
\subsection{Overall Performance}
\vspace{-.2cm}

\begin{table}[t]
\caption{Model performance comparison of segmentation Dice score (normalised to 100\%) on BraTS2018. The best results for each row within tumour types are bolded. ``$\bullet$'' and ``$\circ$'' indicate the availability and absence of the modality for testing, respectively. Last row shows p-value from one-tailed paired t-test.}
\vspace{-1mm}
\setlength\tabcolsep{1pt}
\begin{center}
\resizebox{1.0\linewidth}{!}{
\begin{tabular}{|cccc|cccc|c|cccc|c|cccc|c|}
\hline
\multicolumn{4}{|c|}{Modalities}               & \multicolumn{5}{c|}{Enhancing Tumour}                     & \multicolumn{5}{c|}{Tumour Core}                          & \multicolumn{5}{c|}{Whole Tumour}                         \\ \hline
Fl        & T1        & T1c       & T2        & HMIS  & HVED  & RSeg  & mmFm   & LCKD            & HMIS  & HVED  & RSeg  & mmFm            & LCKD            & HMIS  & HVED  & RSeg  & mmFm   & LCKD            \\ \hline
$\bullet$ & $\circ$   & $\circ$   & $\circ$   & 11.78 & 23.80 & 25.69 & 39.33 & \textbf{45.48}          & 26.06 & 57.90 & 53.57 & 61.21          & \textbf{72.01}          & 52.48 & 84.39 & 85.69 & 86.10 & \textbf{89.45} \\
$\circ$   & $\bullet$ & $\circ$   & $\circ$   & 10.16 & 8.60  & 17.29 & 32.53 & \textbf{43.22} & 37.39 & 33.90 & 47.90 & 56.55          & \textbf{66.58} & 57.62 & 49.51 & 70.11 & 67.52 & \textbf{76.48} \\
$\circ$   & $\circ$   & $\bullet$ & $\circ$   & 62.02 & 57.64 & 67.07 & 72.60 & \textbf{75.65} & 65.29 & 59.59 & 76.83 & 75.41          & \textbf{83.02} & 61.53 & 53.62 & 73.31 & 72.22 & \textbf{77.23} \\
$\circ$   & $\circ$   & $\circ$   & $\bullet$ & 25.63 & 22.82 & 28.97 & 43.05 & \textbf{47.19}          & 57.20 & 54.67 & 57.49 & 64.20          & \textbf{70.17}          & 80.96 & 79.83 & 82.24 & 81.15 & \textbf{84.37} \\
$\bullet$ & $\bullet$ & $\circ$   & $\circ$   & 10.71 & 27.96 & 32.13 & 42.96 & \textbf{48.30}          & 41.12 & 61.14 & 60.68 & 65.91          & \textbf{74.58}          & 64.62 & 85.71 & 88.24 & 87.06 & \textbf{89.97}          \\
$\bullet$ & $\circ$   & $\bullet$ & $\circ$   & 66.10 & 68.36 & 70.30 & 75.07 & \textbf{78.75} & 71.49 & 75.07 & 80.62 & 77.88          & \textbf{85.67} & 68.99 & 85.93 & 88.51 & 87.30 & \textbf{90.47} \\
$\bullet$ & $\circ$   & $\circ$   & $\bullet$ & 30.22 & 32.31 & 33.84 & 47.52 & \textbf{49.01}          & 57.68 & 62.70 & 61.16 & 69.75          & \textbf{75.41}          & 82.95 & 87.58 & 88.28 & 87.59 & \textbf{90.39} \\
$\circ$   & $\bullet$ & $\bullet$ & $\circ$   & 66.22 & 61.11 & 69.06 & 74.04 & \textbf{76.09} & 72.46 & 67.55 & 78.72 & 78.59          & \textbf{82.49} & 68.47 & 64.22 & 77.18 & 74.42 & \textbf{80.10} \\
$\circ$   & $\bullet$ & $\circ$   & $\bullet$ & 32.39 & 24.29 & 32.01 & 44.99 & \textbf{50.09}          & 60.92 & 56.26 & 62.19 & 69.42          & \textbf{72.75}          & 82.41 & 81.56 & 84.78 & 82.20 & \textbf{86.05}          \\
$\circ$   & $\circ$   & $\bullet$ & $\bullet$ & 67.83 & 67.83 & 69.71 & 74.51 & \textbf{76.01} & 76.64 & 73.92 & 80.20 & 78.61          & \textbf{84.85} & 82.48 & 81.32 & 85.19 & 82.99 & \textbf{86.49} \\
$\bullet$ & $\bullet$ & $\bullet$ & $\circ$   & 68.54 & 68.60 & 70.78 & 75.47 & \textbf{77.78} & 76.01 & 77.05 & 81.06 & 79.80          & \textbf{85.24} & 72.31 & 86.72 & 88.73 & 87.33 & \textbf{90.50} \\
$\bullet$ & $\bullet$ & $\circ$   & $\bullet$ & 31.07 & 32.34 & 36.41 & 47.70 & \textbf{49.96}          & 60.32 & 63.14 & 64.38 & 71.52          & \textbf{76.68}          & 83.43 & 88.07 & 88.81 & 87.75 & \textbf{90.46} \\
$\bullet$ & $\circ$   & $\bullet$ & $\bullet$ & 68.72 & 68.93 & 70.88 & 75.67 & \textbf{77.48} & 77.53 & 76.75 & 80.72 & 79.55          & \textbf{85.56} & 83.85 & 88.09 & 89.27 & 88.14 & \textbf{90.90} \\
$\circ$   & $\bullet$ & $\bullet$ & $\bullet$ & 69.92 & 67.75 & 70.10 & 74.75 & \textbf{77.60} & 78.96 & 75.28 & 80.33 & 80.39          & \textbf{84.02} & 83.94 & 82.32 & 86.01 & 82.71 & \textbf{86.73} \\
$\bullet$ & $\bullet$ & $\bullet$ & $\bullet$ & 70.24 & 69.03 & 71.13 & 77.61 & \textbf{79.33} & 79.48 & 77.71 & 80.86 & \textbf{85.78} & 85.31          & 84.74 & 88.46 & 89.45 & 89.64 & \textbf{90.84} \\ \hline
\multicolumn{4}{|c|}{Average}                 & 46.10 & 46.76 & 51.02 & 59.85 & \textbf{63.46}          & 62.57 & 64.84 & 69.78 & 72.97          & \textbf{78.96} & 74.05 & 79.16 & 84.39 & 82.94 & \textbf{86.70} \\ \hline
\multicolumn{4}{|c|}{p-value}                 & 5.3e-6 & 3.8e-7 & 8.6e-7 & 2.8e-5 & -          & 3.7e-5 & 4.1e-7 & 7.2e-6 & 5.3e-7          & - & 1.1e-4 & 1.2e-3 & 1.1e-5 & 5.1e-7 & - \\ \hline
\end{tabular}
}\end{center}
\label{tab:brats2018}
\end{table}

Table \ref{tab:brats2018} shows the overall performance on all 15 possible combinations of missing modalities for three sub-regions of brain tumours. Our models are compared with several strong baseline models: U-HeMIS (abbreviated as HMIS in the figure)~\cite{havaei2016hemis}, U-HVED (HVED)~\cite{dorent2019hetero}, Robust-MSeg (RSeg)~\cite{chen2019robust} and mmFormer (mmFm)~\cite{zhang2022mmformer}. We can clearly observe that with T1c, the model performs considerably better than other modalities for ET. Similarly, T1c for TC and Flair for WT contribute the most, which confirm our motivation.

The LCKD model significantly 
outperforms (as shown by the one-tailed paired t-test for each task between models in the last row of Tab.\ref{tab:brats2018}) U-HeMIS, U-HVED, Robust-MSeg and mmFormer in terms of the segmentation Dice for enhancing tumour and whole tumour on all 15 combinations and the tumour core on 14 out of 15. It is observed that, on average, the proposed LCKD model improves the state-of-the-art performance by 3.61\% for enhancing tumour, 5.99\% for tumour core, and 3.76\% for whole tumour in terms of the segmentation Dice score. Especially in some combinations without the best modality, e.g. ET/TC without T1c and WT without Flair, LCKD has a 6.15\% improvement with only Flair and 10.69\% with only T1 over the second best model for ET segmentation; 10.8\% and 10.03\% improvement with only Flair and T1 for TC; 8.96\% and 5.01\% improvement with only T1 and T1c for WT, respectively. These results demonstrate that useful knowledge of the best modality has been successfully distilled into the model by LCKD for multimodal learning with missing modalities.

\vspace{-.2cm}
\subsection{Analyses}
\vspace{-.2cm}

\begin{table}[t]
\caption{Different LCKD variants Dice score. LCKD-s and LCKD-m represent LCKD with single teacher and multi-teacher, respectively.}
\begin{center}
\resizebox{0.57\linewidth}{!}{
\begin{tabular}{|cccc|cc|cc|cc|}
\hline
\multicolumn{4}{|c|}{Modalities}               & \multicolumn{2}{c|}{Enhancing Tumour}                     & \multicolumn{2}{c|}{Tumour Core}                          & \multicolumn{2}{c|}{Whole Tumour}                         \\ \hline
Fl        & T1        & T1c       & T2        & LCKD-s          & LCKD-m          & LCKD-s          & LCKD-m          & LCKD-s          & LCKD-m          \\ \hline
$\bullet$ & $\circ$   & $\circ$   & $\circ$   & \textbf{46.19} & 45.48          & \textbf{72.51} & 72.01          & 89.38          & \textbf{89.45} \\
$\circ$   & $\bullet$ & $\circ$   & $\circ$   & 43.05          & \textbf{43.22} & 65.79          & \textbf{66.58} & 75.86          & \textbf{76.48} \\
$\circ$   & $\circ$   & $\bullet$ & $\circ$   & 74.26          & \textbf{75.65} & 81.93          & \textbf{83.02} & 77.14          & \textbf{77.23} \\
$\circ$   & $\circ$   & $\circ$   & $\bullet$ & \textbf{48.59} & 47.19          & \textbf{70.64} & 70.17          & 84.25          & \textbf{84.37} \\
$\bullet$ & $\bullet$ & $\circ$   & $\circ$   & \textbf{49.65} & 48.30          & \textbf{74.98} & 74.58          & \textbf{90.12} & 89.97          \\
$\bullet$ & $\circ$   & $\bullet$ & $\circ$   & 77.72          & \textbf{78.75} & 85.37          & \textbf{85.67} & 90.33          & \textbf{90.47} \\
$\bullet$ & $\circ$   & $\circ$   & $\bullet$ & \textbf{49.86} & 49.01          & \textbf{75.65} & 75.41          & 90.28          & \textbf{90.39} \\
$\circ$   & $\bullet$ & $\bullet$ & $\circ$   & 75.32          & \textbf{76.09} & 81.77          & \textbf{82.49} & 79.96          & \textbf{80.10} \\
$\circ$   & $\bullet$ & $\circ$   & $\bullet$ & \textbf{51.65} & 50.09          & \textbf{73.95} & 72.75          & \textbf{86.39} & 86.05          \\
$\circ$   & $\circ$   & $\bullet$ & $\bullet$ & 75.34          & \textbf{76.01} & 84.21          & \textbf{84.85} & 86.05          & \textbf{86.49} \\
$\bullet$ & $\bullet$ & $\bullet$ & $\circ$   & 77.42          & \textbf{77.78} & 84.79          & \textbf{85.24} & \textbf{90.50} & \textbf{90.50} \\
$\bullet$ & $\bullet$ & $\circ$   & $\bullet$ & \textbf{51.05} & 49.96          & \textbf{76.84} & 76.68          & 90.39          & \textbf{90.46} \\
$\bullet$ & $\circ$   & $\bullet$ & $\bullet$ & 76.97          & \textbf{77.48} & 84.93          & \textbf{85.56} & 90.83          & \textbf{90.90} \\
$\circ$   & $\bullet$ & $\bullet$ & $\bullet$ & 77.53          & \textbf{77.60} & 83.95          & \textbf{84.02} & 86.71          & \textbf{86.73} \\
$\bullet$ & $\bullet$ & $\bullet$ & $\bullet$ & 78.39          & \textbf{79.33} & 85.26          & 85.31          & 90.74          & \textbf{90.84} \\ \hline
\multicolumn{4}{|c|}{Average}                 & \textbf{63.53} & 63.46          & 78.84          & \textbf{78.96} & 86.60          & \textbf{86.70} \\ \hline
\multicolumn{4}{|c|}{Best Teacher}  & \multicolumn{2}{c|}{T1c}              & \multicolumn{2}{c|}{T1c}         & \multicolumn{2}{c|}{Fl}  \\ \hline
\end{tabular}
}\end{center}
\label{tab:variants}
\end{table}

\subsubsection{Single Teacher vs. Multi-Teacher}

To analyse the effectiveness of knowledge distillation from multiple teachers of all tasks in the proposed LCKD model, we perform a study to compare the model performance of adopting single teacher and multi-teachers for knowledge distillation. We enable multi-teachers for LCKD by default to encourage the model parameters to move to a point that can perform well for all tasks. However, for single teacher, we modify the function $\phi(.)$ in Eq.~\ref{eq:teacher_elect2} to pick the modality with max appearance time (e.g. if we have \{T1c, T1c, Flair\} then $\phi(.)$ returns \{T1c\}), while keeping other settings the same.

\begin{figure}[]
\begin{minipage}{.63\linewidth}
\centering
\includegraphics[width=1.0\textwidth]{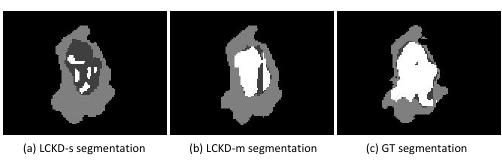}
\vspace{-3mm}
\caption{Segmentation Visualisation with only T2 input available. Light grey, dark grey and white represent different tumour sub-regions.}
\label{fig:seg-vis}
\end{minipage}
\begin{minipage}{.35\linewidth}
\centering
\includegraphics[width=1.0\textwidth]{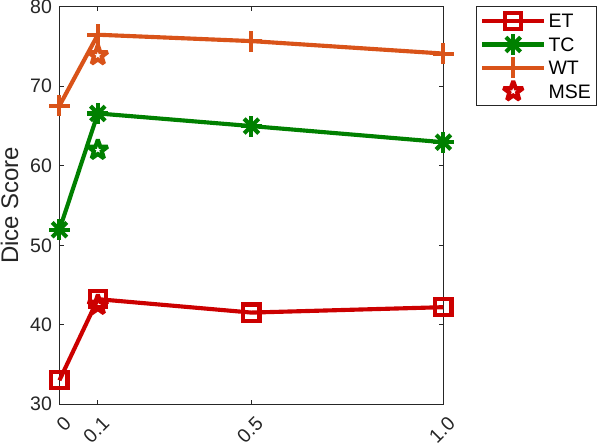}
\vspace{-3mm}
\caption{T1 Dice score as function of $\alpha$ (L1 loss for $\ell_{ckd}(.)$ in (\ref{eq:kd})). Star markers show the Dice score for L2 loss for $\ell_{ckd}(.)$ ($\alpha=.1$). Colors denote different tumors.}
\label{fig:sen}
\end{minipage}
\end{figure}

From Table \ref{tab:variants}, compared with multi-teacher model LCKD-m, we found that the single teacher model LCKD-s receives comparable results for ET and TC segmentation (it even has better average performance on ET), but it cannot outperform LCKD-m on WT. This phenomenon, also shown in Fig.\ref{fig:seg-vis}, demonstrates that LCKD-m has better overall segmentation performance. This resonates with our expectations because there are 3 tasks in BraTS, and the best teachers for ET and TC are the same, which is T1c, but for WT, Flair is the best one. Therefore, for LCKD-s, the knowledge of the best teacher for ET and TC can be distilled into the model, but not for WT. The LCKD-m model can overcome this issue since it attempts to find a point in the parameter space that is beneficial for all tasks. Empirically, we observed that both models found the correct teacher(s) quickly: the best teacher of the single teacher model alternated between T1c and Flair for a few validation rounds and stabilised at T1c; while the multi-teacher model found the best teachers (T1c and Flair) from the first validation round.

\subsubsection{Role of $\alpha$ and CKD Loss Function}
As shown in Fig.~\ref{fig:sen}, we set  $\alpha$ in Eq.\ref{eq:tot_loss} to 
$\{0,0.1,0.5,1\}$ using T1 input only and L1 loss for $\ell_{ckd}(.)$ in (\ref{eq:kd}). 
If $\alpha=0$, the model performance drops greatly, but when $\alpha > 0$, results improve, where $\alpha=0.1$ produces the best result. This shows the importance of the cross-modal knowledge distillation loss in~(\ref{eq:tot_loss}).
To study the effect of a different CKD loss, we show Dice score with L2 loss for $\ell_{ckd}(.)$ in (\ref{eq:kd}), with $\alpha=0.1$.
Compared with the L1 loss, we note that Dice decreases slightly with the L2 loss, especially for TC and WT.

\vspace{-.2cm}
\section{Conclusion}
\vspace{-.2cm}

In this paper we introduced the Learnable
Cross-modal Knowledge Distillation (LCKD), which is the first method that can handle missing modality during training and testing by distilling knowledge from automatically selected important modalities for all training tasks to train other modalities.
Experiments on BraTS2018~\cite{menze2014multimodal,bakas2018identifying} show that LCKD reaches  state-of-the-art performance in missing modality segmentation problems.
We believe that our proposed LCKD has the potential to allow the use of multimodal data for training and missing-modality data per testing.
One point to improve about LCKD is the greedy teacher selection per task. We plan to improve this point by transforming this problem into a meta-learning strategy, where the meta parameter is the weight for each modality, which will be optimised per task.

%
%
\bibliographystyle{splncs04}
\bibliography{mybib}

\section*{Appendix}

\section*{Segmentation Visualisations}

\begin{figure}[htbp]
\centering
\includegraphics[width=1\linewidth]{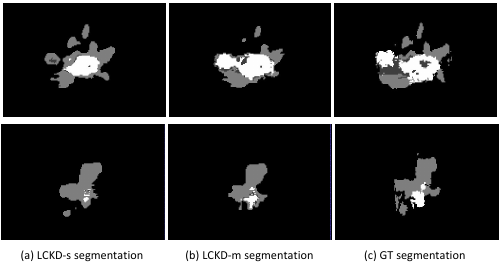}
\caption{More segmentation visualisations with only T2 input available. The two rows represent two case studies and columns denote segmentation with single (1st column) or multiple (2nd column) teachers, and from ground truth (3rd column). Light grey, dark grey and white represent different tumour sub-regions.} 
\label{fig:seg-vis-supp}
\end{figure}

\newpage

\section*{Model translates to other domains}

We also adapted our method and obtained results for the multi-modal analysis with missing modality for the Heart Segmentation (MMWHS) problem by training with random dropping modalities for 4000 epochs. Evaluated on the CT modality only, compared with a baseline model (a multi-modal model that replaces missing modality inputs with 0s), our LCKD model improves Dice from 90.7 to 92.2 on the left ventricle (LV) and from 86.1 to 88.0 on the myocardium (Myo). Using the MR modality only, the improvements are more obvious: from 78.6 to 84.7 on LV, and from 62.8 to 68.4 on Myo. The aforementioned results demonstrate the effectiveness of the LCKD model on other domains.

\section*{Percentage of Modalities Selected as Teachers}

\begin{figure}[htbp]
\centering
\includegraphics[width=0.6\linewidth]{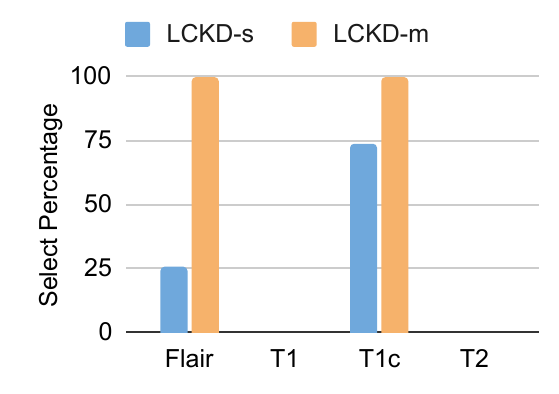}
\caption{Percentage of different modalities selected as teachers for LCKD-s and LCKD-m models. We perform the teacher election procedure every 5000 iterations. T1 and T2 are not selected as teachers in the whole process, so they have been selected 0\% of time.}
\label{fig:select-percent}
\end{figure}

\end{document}